\documentclass[conference]{IEEEtran}
\IEEEoverridecommandlockouts
\usepackage{cite}
\usepackage{amsmath,amssymb,amsfonts}
\usepackage{algorithmic}
\usepackage{graphicx}
\usepackage{textcomp}
\usepackage{xcolor}
\usepackage[utf8]{inputenc}
\usepackage{booktabs}
\usepackage{comment}
\usepackage[hidelinks]{hyperref}

\def\BibTeX{{\rm B\kern-.05em{\sc i\kern-.025em b}\kern-.08em
    T\kern-.1667em\lower.7ex\hbox{E}\kern-.125emX}}
\begin{document}

\title{GATNextHop: A GAT for Shortest Path Routing with Cross-Topology Generalization\thanks{Code available at \url{https://github.com/knhn1004/GATNextHop}}}

\author{\IEEEauthorblockN{Chia-Hong Chou}
\IEEEauthorblockA{\textit{Department of Computer Science} \\
\textit{San José State University}\\
San José, CA \\
oliver.chou@sjsu.edu}
\and
\IEEEauthorblockN{Katerina Potika}
\IEEEauthorblockA{\textit{Department of Computer Science} \\
\textit{San José State University}\\
San José, CA \\
katerina.potika@sjsu.edu}
}

\maketitle

\begin{abstract}
Common shortest-path algorithms, such as Dijkstra’s (SPF), that OSPF uses, provide exact routing solutions but must be recomputed for each network topology, limiting scalability in dynamic or large-scale networks. This paper proposes the GATNextHop model to determine whether a Graph Neural Network, namely the Graph Attention Network, can approximate shortest paths and generalize across topologies. By training on synthetic graphs and evaluating on real-world Internet Service Provider networks from the Internet Topology Zoo, we aim to benchmark our model’s ability to learn routing heuristics that transfer across network structures. Performance will be evaluated in terms of accuracy, inference speed, and generalization, comparing the GNN against Dijkstra’s algorithm to quantify trade-offs between learned and classical routing approaches.
\end{abstract}

\begin{IEEEkeywords}
Graph Neural Networks, shortest path, routing, topology, Internet Topology Zoo, Graph Attention Networks, Random Graphs
\end{IEEEkeywords}

\section{Introduction}
In modern network environments where the topology constantly changes, it is challenging to efficiently route packets in an optimal manner. Although shortest-path routing protocols such as OSPF \cite{moy1997ospf} and its SPF (Dijkstra's) algorithm are reliable and exact, they must be rerun per topology, insufficient in network topologies with frequent topology changes or limited transmissions. 

Graph Neural Networks (GNNs) introduced a paradigm shift in existing routing optimization studies. Classical algorithms, including Dijkstra's, guarantee an exact solution that operates optimally in static topologies but struggle to adapt to scalable, dynamic environments.
In contrast, according to Jiang et al.~\cite{Jiang2024Graph}, the GNN-based approach enables optimization in changing environments and has been applied to solve network routing problems.

Previous work explored using GNNs to perform routing optimizations in dynamic, scalable network topologies. It was discovered that GNN-based routing methods could outperform Dijkstra's shortest path algorithm in dynamic or complex network environments~\cite{Mary2025Graph-Based}.
Following this idea, we investigate whether a GNN, specifically a Graph Attention Network (GAT) \cite{Velickovic2017Graph}, can approximate the most likely next-hop node and generalize from synthetic (random) graphs to real internet service provider (ISP) topologies.

First, we analyze real-world network topologies from the Internet Topology Zoo \cite{knight2011internet} dataset and explore key characteristics at the node and edge-level, including degree, centrality, weights, betweenness, and clustering coefficient.
Then, based on the network analysis of real-world networks, we curated synthetic (random) datasets of $1000$ graphs using the NetworkX library \cite{Hagberg2008Exploring}, including features similar to those in the training set.
Using the training set, we performed an $80-20$ test--validation split and trained a GAT that encodes edge and node features to predict the best next-hop node for the shortest distance. Using GAT with $4$ node features (normalized degree, betweenness centrality, clustering coefficient, and degree centrality) and $1$ edge feature (edge weight), we achieved $85.1\%$ accuracy in synthetic validation during training and $84.2\%$ accuracy in the Internet Topology Zoo test set.
Additionally, by including only betweenness centrality as the node-level feature, the model outperformed the full-feature model and achieved $85.7\%$ accuracy on the synthetic set and $84.6\%$ on the test set.
It shows that GAT has the potential to generalize from a synthetic data set and estimate the optimal next-hop on real-world network topologies, provided structurally similar training data.

The paper's contribution is the following.
(i) Designed GATNextHop, a GAT-based next-hop prediction model achieving $85.1\%$ accuracy on synthetic graphs and $84.2\%$ on unseen Internet Topology Zoo topologies, with ablation studies identifying betweenness centrality as the most impactful node feature.
(ii) Benchmarked SPF (Dijkstra) vs. GAT inference speed across graph sizes using $180$ real-world ISP topologies from the Internet Topology Zoo dataset.

The goal of our approach is to handle conditions in which Dijkstra's assumptions break down (e.g., dynamic and partial graphs). Our focus is on generalizing under uncertainty rather than computational competition.

\section{Related Work}

Almasan et al.~\cite{Almasan2019Deep} combined GNNs with deep reinforcement learning (DRL) to learn routing policies that generalize to unseen topologies, outperforming previous DRL and MLP approaches.
Rusek et al.~\cite{rusek2020routenet} introduced RouteNet, which uses GNN to predict per-path delay and loss, and applied it in software-defined networking (SDN) routing optimization.
Ferriol-Galmés et al.~\cite{Ferriol-Galmes2022RouteNet-Fermi} took a step further and developed RouteNet-Fermi, which is more accurate and works on larger networks unseen during training.
He et al.~\cite{He2024Routing} proposed MPDRL, which plugs a GNN into the DRL agent to exploit message transmission across the topology and improve load-balanced routing compared to traditional algorithms in ISP networks.
Zheng et al.~\cite{Zheng2022Research} presented GNN-DRL for SDN, using a GNN state encoder with DRL to minimize maximum link utilization and delay that outperforms OSPF, ECMP, and the previous intelligent router (EARS) under high load.

However, most previous work uses vanilla or modified graph convolutional networks (GCNs)~\cite{Kipf2016Semi-Supervised} or message-passing neural networks (MPNNs)~\cite{Gilmer2017Neural} with uniform or degree-normalized weights over neighbors. On the other hand, the graph attention network (GAT)~\cite{Velickovic2017Graph} introduces masked self-attention, allowing different importance assignments for each neighbor.
For our algorithm, it is reasonable to use a GAT to solve the next-hop problem because our goal is to encode structural information and predict patterns. This aligns with GAT's philosophy of treating each neighbor differently, thereby providing additional learning opportunities for better decision-making.
Admittedly, our use case is limited to static environments and does not analyze dynamic topology configurations. Nevertheless, the paper explores whether GAT can learn from seen topologies and generalize to predicting routes in unseen topologies.

\section{Methodology}
\subsection{Real-World Network Topology Analysis}
We target the real-world Internet Topology Zoo dataset (Zoo) \cite{knight2011internet} ($276$ raw topologies and $180$ after pre-processing).
The topology analysis follows the following steps: (1) pre-process the raw data, (2) assign weights on graphs, (3) compute distance matrices, (4) profile the graph statistics.

\subsubsection{Pre-processing}
We convert each graph to an undirected simple graph (removing multi-edges and self-loops), extract the largest connected component, and re-label nodes $0\ldots n$. Graphs with fewer than three nodes or parser failures were discarded, leaving $180$ graphs.

\subsubsection{Weight Assignment}
Since real graphs lack edge weights, we assign each edge a uniform random weight in $[1, \ldots, 100]$ of real numbers with 64-bit precision.

\subsubsection{Distance Matrices}
We compute all-pairs shortest paths via Dijkstra and store them as ground truth. Next-hop labels are derived during evaluation: for each $(s,t)$, select the neighbor $w$ of $s$ satisfying $dist(s,t) = weight(s,w) + dist(w,t)$, where $weight(e)$ is the weight of edge $(e)$, and $dist(s,t)$ is the weight of the shortest path from node $s$ to node $t$.

\subsubsection{Graph Profiling}
\label{sec:methodology:pre-processing:graph-profiling}
We collect statistics of the $180$ processed Internet Topology Zoo graphs and take the medians
as shown in Table~\ref{tab:zoo-profile}. This shapes the target parameters we want when generating the synthetic training graphs.

\begin{table}[htbp]
\centering
\caption{Summary statistics of 180 Internet Topology Zoo graphs.}
\label{tab:zoo-profile}
\begin{tabular}{lrrrr}
\toprule
Metric & Median & Min & Max & Std \\
\midrule
Node count       &  25.00 &   4.00 & 158.00 & 25.09 \\
Edge count       &  31.00 &   4.00 & 193.00 & 30.10 \\
Density          &   0.09 &   0.02 &   1.00 &  0.13 \\
Avg.\ degree     &   2.25 &   1.60 &   8.00 &  0.63 \\
Clustering       &   0.03 &   0.00 &   1.00 &  0.15 \\
Avg.\ shortest path & 3.17 &   1.00 &  13.07 &  2.06 \\
Assortativity    &  $-0.35$ & $-1.00$ &   0.32 &  0.27 \\
Diameter         &   7.00 &   1.00 &  35.00 &  5.56 \\
\bottomrule
\end{tabular}
\end{table}

\subsection{Synthetic Dataset Curation}
Based on the graph profiling, we generated $1{,}000$ synthetic graphs ($200$ each) from five models: Erdős–Rényi (ER), Barabási–Albert (BA), Watts–Strogatz (WS), Stochastic Block Model (SBM), and Waxman, calibrated to the Zoo medians.


The parameters were cherry-picked to make the synthetic data resemble the Zoo set as closely as possible, following the results of earlier graph profiling, these parameters are:
\begin{itemize}
    \item ER: $p$ chosen so that expected degree = 2.3
    \item BA: $m=1$ so mean degree $\approx2$.
    \item WS: $k=2,p\in[0.1,0.4]$
    \item SBM: internal target degree $2.3$, low p\_out
    \item Waxman: lower $\alpha/\beta$ for sparser graphs
\end{itemize}

Additionally, the all-pairs shortest path values ($dist(u,v)$)
are precomputed and the original $1,000$ graphs were split into $800$ train and $200$ validation using a random seed of $42$.

\subsection{GNN Model Design}
\label{sec:GNN-model-design}

\subsubsection{Task formulation and Ground Truth}
Formally, our goal is that given a graph $G$, a source node $s$, and a destination $t$, we want to predict which neighbor $w$ of $s$ lies on the shortest path between $s$ and $t$, essentially the next-hop node. This is a classification problem over the candidate set $N(s)$. The ground truth was a neighbor $w$, where $dist(s, t) = weight(s, w) + dist(w, t)$ verified with a tolerance of $\text{1e-5}$.

We designed a Graph Attention Network (GAT)~\cite{Velickovic2017Graph} called GATNextHop model. We chose GAT for its attention mechanism as it can upweight neighbors that are ``toward" the shortest paths and downweight others. A GAT fits well in this neighbor-dependent adaptive decision-making problem.

\subsubsection{Feature Selection}
To capture the structural information, we use 4-dimensional node features, each normalized to $[0, 1]$ by dividing by the maximum value: degree, betweenness centrality (weighted), clustering coefficient, and degree centrality. {Degree counts a node's direct connections, betweenness centrality measures how often a node lies on shortest paths between other pairs, clustering coefficient captures the density of triangles around a node, and degree centrality is the normalized degree. Additionally, it uses a $1$-dimensional edge feature consisting of the edge weight.

\subsubsection{Architecture (GATNextHop)}
\begin{figure}[h]
    \centering
    \includegraphics[width=1\linewidth]{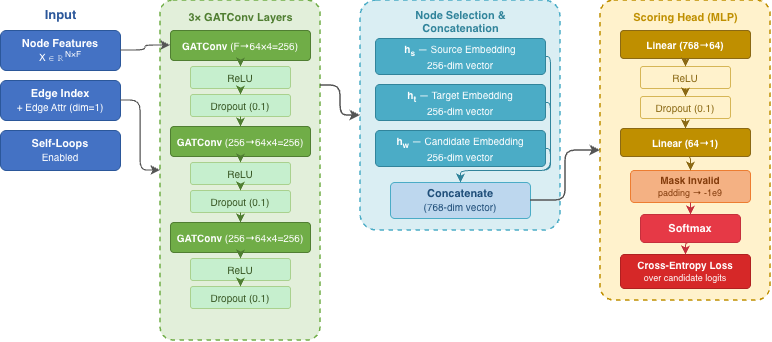}
    \caption{GATNextHop GAT Model Architecture}
    \label{fig:model-architecture}
\end{figure}
As shown in Fig~\ref{fig:model-architecture}.
\begin{itemize}
    \item GATConv 3 layers, each with $64$ hidden units and $4$ attention heads (output dim = $64 \times 4 = 256$ per layer). Edge-aware attention using \texttt{edge\_dim=1}. Self-loops enabled. ReLU activation + dropout (0.1) after each layer.
    \item For each $(s, t, \text{candidate } w)$ triple, concatenate the final embeddings $[h_s; h_t; h_w]$ ($dim = 3 \times 256 = 768$). Pass through a 2-layer MLP: Linear(768, 64) $\to$ ReLU $\to$ Dropout(0.1) $\to$ Linear(64, 1) to produce a scalar score per candidate.
    \item Invalid candidates (padding) masked to $-1e9$ before softmax.
    \item Loss function: Cross-entropy over candidate logits.
\end{itemize}

\subsubsection{Training Details}
We used an Adam optimizer, with initial learning rate $LR=1e-3$. A ReduceLROnPlateau learning rate scheduler is used that monitors validation accuracy by a factor of $0.5$ and patience $5$. We ran $100$ epochs (with early stopping) training with a batch size of $16$ graphs, and sampled $100$ $(s,t)$ pairs per graph per epoch, and there are a maximum of $64$ candidates per source node. For the results shown, the model was trained on a MacBook Pro M2 Max device with a CPU and took $\text{3 min 27 s}$, early stopping at the 77th epoch.

\subsection{Experiments}

\subsubsection{Gap Analysis: Synthetic vs Internet Topology Zoo Graphs}
Despite our best effort to create a synthetic data set that resembles the characteristics of the Internet Topology Zoo, the synthetic graph generated uses distinct models containing varied attributes. Recognizing this gap, we compare 8 structural metrics (node count, edge count, density, average degree, clustering, average shortest path, assortativity, diameter) between Zoo and synthetic using median, ratio, and a realism score.
Ratio is calculated as $\text{(synthetic median)}/\text{(Zoo median)}$ for each metric with $\text{Zoo median}\neq0$. The Realism score for each metric is computed as $1-D, D\in[0.0,1.0]$ where $D$ is the Kolmogorov-Smirnov statistic distance  \cite{Massey1951The} between the Zoo and the synthetic distributions, so a $1.0$ will represent an identical distribution and $0.0$ shows a maximum discrepancy.

\subsubsection{Main Evaluation}
We train the GAT on 800 synthetic graphs and validate on the other 200 graphs. The model never sees the actual Internet Topology Zoo graph during training. The accuracy is reported as $\text{fraction of (s, t) pairs}$  where the predicted next-hop matches ground truth.

\subsubsection{Ablation Study}
To study what contributes to next-hop prediction, we conducted an ablation study in which only each of the four node features was included, i.e., degree-only, betweenness-only, clustering-coefficient-only, and degree-centrality-only.
This gives us a clearer picture of how each node-level feature contributes to next-hop prediction.

\subsubsection{SPF vs GAT benchmark}
We also want to see if GAT has the potential to be faster than the shortest path first algorithm (SPF) used by OSPF, so we compare Dijkstra's (analogous to SPF) wall-clock time vs GAT's single query inference time.
We used 200 $\text{(s, t)}$ pairs per graph and reported median timing by graph size bucket (small $<50$ nodes, medium $50-150$, and large $\geq150$) with 3 warm-up rounds before timing.

\section{Results \& Discussion}

\subsection{Synthetic vs Internet Topology Zoo Graph Comparison}
Figure~\ref{fig:radar_comparison} compares each synthetic model with the Zoo median (normalized to~$1.0$).
The SBM has a clustering coefficient five times the Zoo median, which means it is more clustered than the real-world topology.
Waxman is high in density (around $3.5$ times the Zoo median), which shows that it is denser than the real set. The WS model over-represents the diameter and average shortest path, representing a sparser training set.

\begin{figure}[htbp]
    \centering
    \includegraphics[width=0.8\linewidth]{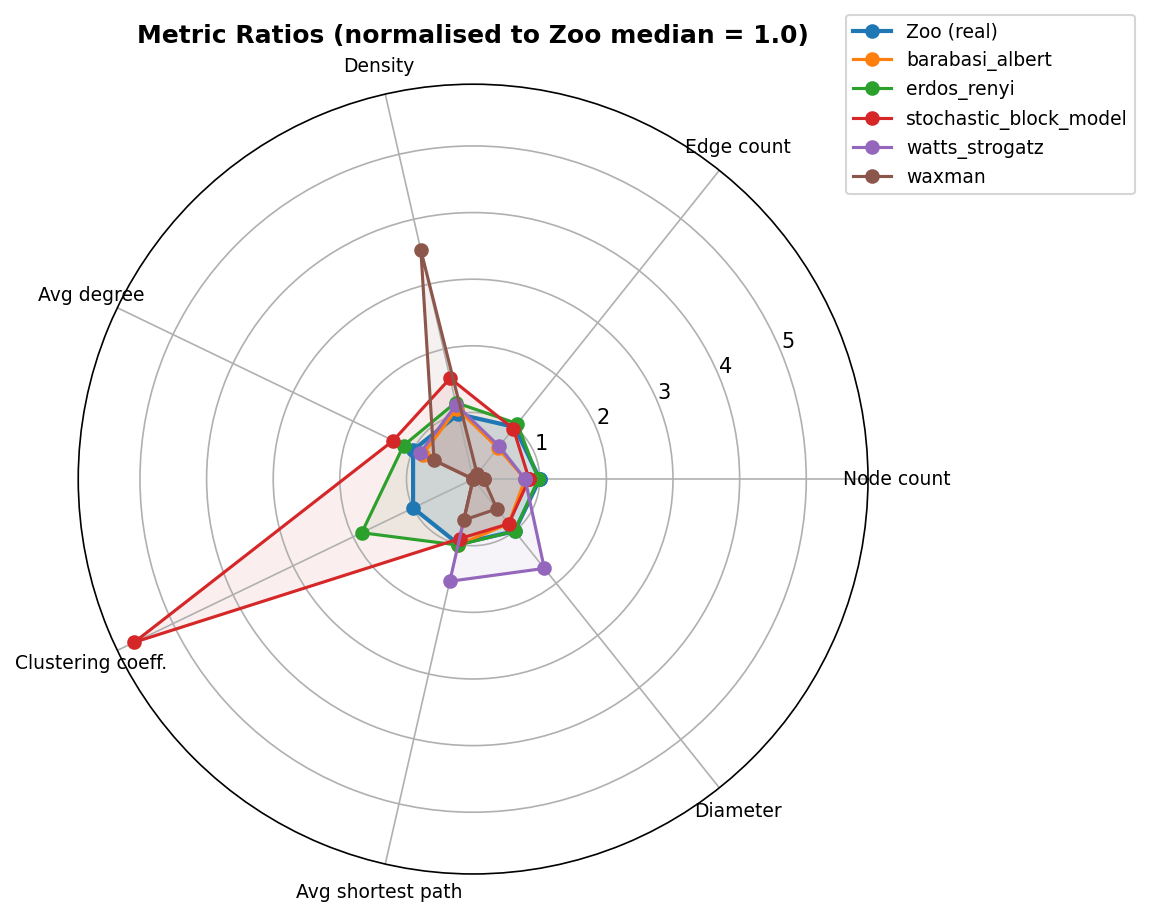}
    \caption{Radar comparison of Internet Topology Zoo and synthetic graphs}
    \label{fig:radar_comparison}
\end{figure}

The box plots in Figure~\ref{fig:comparison_boxplots} show the overall comparison of the synthetic versus realistic Zoo set in the six different metrics: node count, edge count, density, average degree, clustering coefficient, and average shortest path. Note that the medians are similar, but the outliers on synthetic vs Zoo are different, which is ideal because we want to test the generalization capability from synthetic training to testing on real data.

\begin{figure}[htbp]
    \centering
    \includegraphics[width=\linewidth]{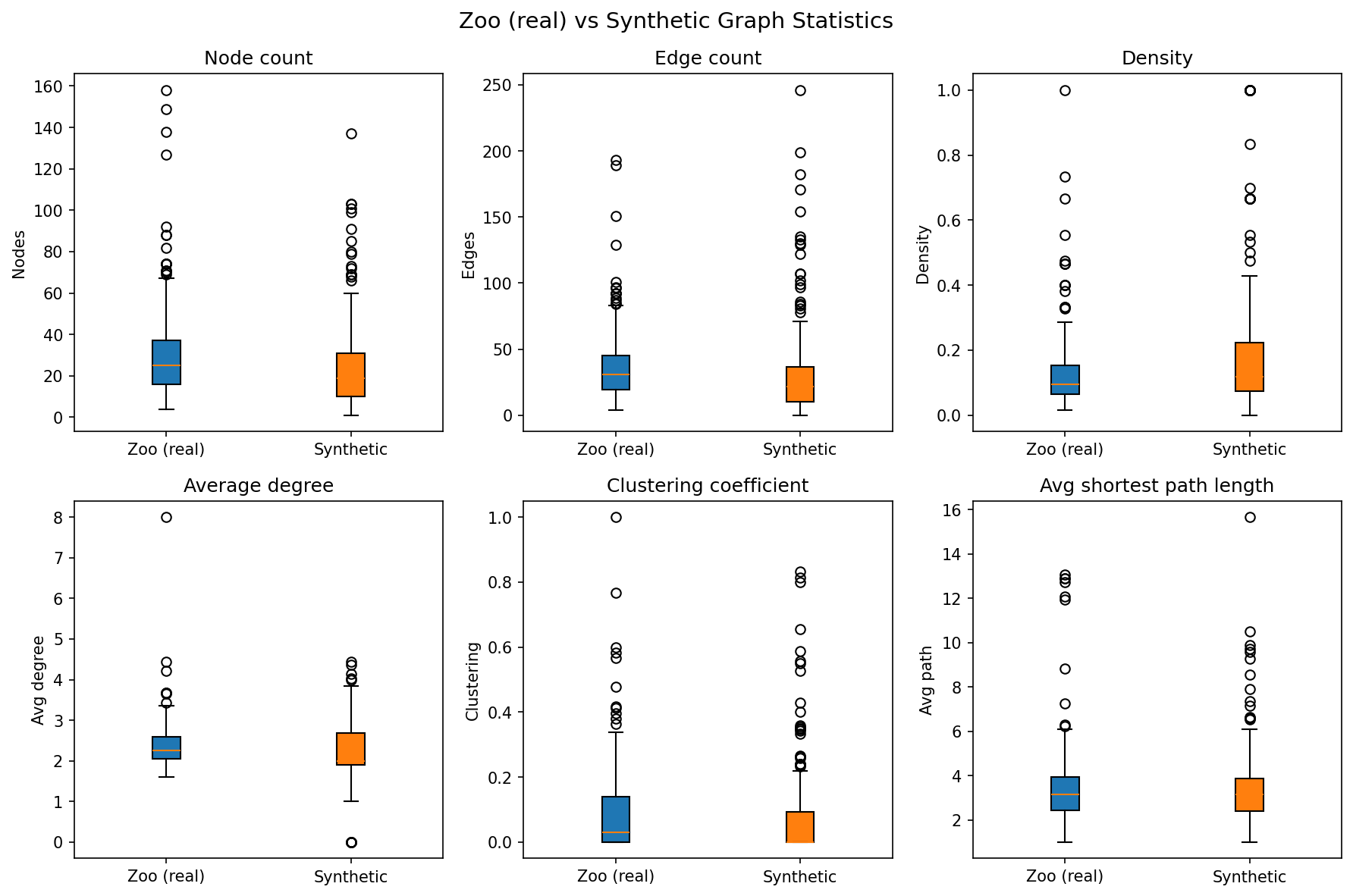}
    \caption{Boxplots of graph statistics by dataset (Zoo vs synthetic) for each metric.}
    \label{fig:comparison_boxplots}
\end{figure}

Table~\ref{tab:gap_analysis} shows the medians, ratios, and Realism computations of the synthetic graph versus the Zoo real set. The diameter is the most-matched metric with a $1.00$ ratio and $0.96$ realism. The average shortest path also matches well with a $1.00$ ratio and $0.93$ realism. This is significant because distance affects the prediction of the shortest path in a large way. Density, node count, and clustering coefficient are well-matched metrics, whereas edge count, assortativity, and average degree show a larger mismatch. These discrepancies and the lower realism may explain why in ablation studies, the clustering coefficient may be a weaker node-level feature. If a data set with better realism across these metrics were used, the ablation studies may reveal stronger results when different node-level features were used.

\begin{table}[htbp]
\centering
\caption{Zoo vs synthetic graph statistics (median and ratio).}
\label{tab:gap_analysis}
\begin{tabular}{lrrrr}
\toprule
Metric & Zoo median & Syn.\ median & Ratio & Realism \\
\midrule
Node count & 25.00 & 19.00 & 0.76 & 0.81 \\
Edge count & 31.00 & 21.50 & 0.69 & 0.78 \\
Density & 0.09 & 0.12 & 1.24 & 0.85 \\
Avg degree & 2.25 & 2.00 & 0.89 & 0.65 \\
Clustering coeff. & 0.03 & 0.00 & --- & 0.83 \\
Avg shortest path & 3.17 & 3.16 & 1.00 & 0.93 \\
Assortativity & -0.35 & -0.22 & 0.63 & 0.74 \\
Diameter & 7.00 & 7.00 & 1.00 & 0.96 \\
\bottomrule
\end{tabular}
\end{table}

\subsection{Model Evaluation}

Table~\ref{tab:results} shows the training result of our GATNextHop model. The training plateaued and early stopped at epoch $78$. Both validation training and validation loss steadily decrease throughout the training. In particular, the validation loss is consistently lower than that of training, and the validation accuracy is also consistently higher. The validation loss is consistently lower than the training loss due to regularization, and the curves show no signs of overfitting.

\begin{figure}[htbp]
    \centering
    \includegraphics[width=\linewidth]{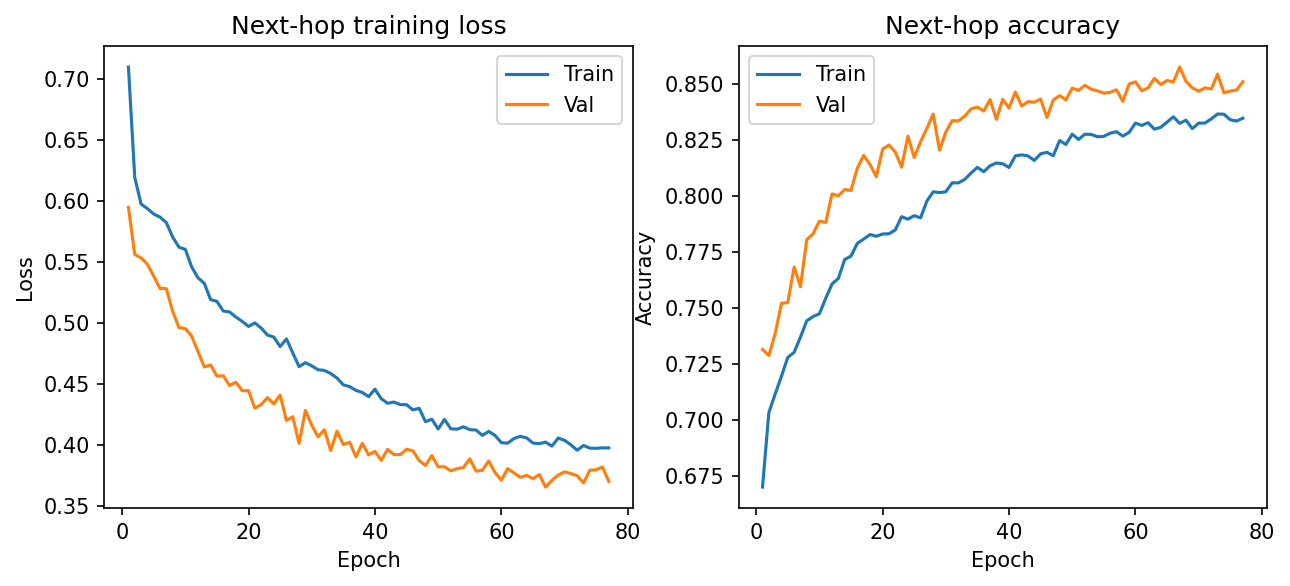}
    \caption{Training and validation loss (left) and accuracy (right) early stopped at epoch 78 on synthetic graphs. Validation accuracy surpasses training accuracy throughout, indicating strong generalization within the synthetic distribution.}
    \label{fig:training_curves}
\end{figure}

Table~\ref{tab:results} shows the final result of the model, comparing synthetic validation (Syn. Val) and testing against the Internet Topology Zoo test set (Zoo Test). The key takeaway is the small gap between the synthetic validation ($85.1\%$) and the Zoo Test result ($84.2\%$). This shows that our approach, using a synthetic dataset for training, generalizes well and confirms the effectiveness of earlier synthetic data curation strategies.

\begin{table}[htbp]
\centering
\caption{Next-hop classification results. Synthetic vs Zoo.}
\label{tab:results}
\begin{tabular}{lcc}
\toprule
Metric & Synthetic\ Validation & Zoo Testing \\
\midrule
Accuracy & 0.851 & 0.842 \\
Number of graphs & 200 & 180 \\
\bottomrule
\end{tabular}
\end{table}

\subsection{Ablation Study}
Table~\ref{tab:ablation} compares models trained on each node-level feature alone and on all four combined. Interestingly, betweenness centrality is the single most helpful feature for predicting the next-hop. This is expected because betweenness captures how often a node lies on shortest paths, which matches our next-hop definition. Degree and degree centrality perform similarly because degree centrality is simply degree normalized by $(n{-}1)$. For graphs of similar size, both convey the same structural signal. The clustering coefficient alone is a weak feature, since it reflects local triangle density and is less relevant to shortest-path choice. Ablating the other three features from the full model, i.e., using betweenness only, improves test performance ($84.6\%$ vs $84.2\%$ in the Zoo test), which suggests that betweenness alone captures the structure relevant to next-hop prediction and that degree, degree centrality, and clustering add little or no benefit and may introduce noise.
\begin{table}[htbp]
\centering
\caption{Ablation: effect of individual node features on next-hop classification.}
\label{tab:ablation}
\begin{tabular}{@{}lcccc@{}}
\toprule
Features & Epoch Stopped & Syn.\ Val & Zoo Test & Zoo Loss \\
\midrule
Clustering only & 37 & 0.649 & 0.709 & 0.569 \\
Degree only & 55 & 0.816 & 0.811 & 0.446 \\
Degree centr.\ only & 78 & 0.824 & 0.818 & 0.444 \\
Betweenness only & 75 & \textbf{0.857} & \textbf{0.846} & \textbf{0.379} \\
\midrule
All four & 77 & 0.851 & 0.842 & 0.386 \\
\bottomrule
\end{tabular}
\end{table}

\subsection{SPF vs GAT benchmark}
Testing on a M2 Max MacBook Pro device with CPU, we compared the single-source Dijkstra (SPF) wall-clock time vs GAT single-query inference time over 180 Internet Topology Zoo graphs, and the results shown in Table~\ref{tab:benchmark} are grouped by size. Across all graphs, SPF takes a median of $0.01$ ms, whereas GAT takes $0.61$ ms. However, the gap becomes narrower as node size grows larger when the speed up goes from $0.02$ in small $n<50$ bucket to $0.03$ in $n\ge150$ bucket. This trend corresponds to the constant overhead of GAT as a bottleneck at small $n$, while Dijkstra's $O((V+E)\log V) $ cost increases.

Figure~\ref{fig:spf_vs_gat} visualizes this trend.
Speedup is defined as SPF time / GAT time. Values below 1.0 indicate GAT is slower. The narrowing gap from 0.02 to 0.03 as n grows shows Dijkstra's linear scaling versus GAT's constant overhead.

Even though our setup shows a degradation in speed performance with GAT for a single query, it suggests that we can potentially benefit from GNNs in larger graphs. Additionally, GAT offers fixed-cost inference regardless of the number of queries on the same topology, where each additional pairwise query requires only another MLP forward pass, amortizing the cost across large queries.

\begin{table}[htbp]
\centering
\caption{SPF (Dijkstra) vs GAT next-hop inference on Topology Zoo.}
\label{tab:benchmark}
\begin{tabular}{lrrr}
\toprule
 & SPF (ms) & GAT (ms) & Speedup \\
\midrule
All graphs (n=180) & 0.01 & 0.61 & 0.02 \\
\midrule
Small ($n<50$) & 0.01 & 0.58 & 0.02 \\

Medium ($50\leq n<150$) & 0.04 & 1.43 & 0.02 \\

Large ($n\geq150$) & 0.08 & 2.54 & 0.03 \\

\midrule
GAT accuracy (\%) & --- & 84.6 & --- \\
\bottomrule
\end{tabular}
\end{table}

\begin{figure}[htbp]
    \centering
    \includegraphics[width=\linewidth]{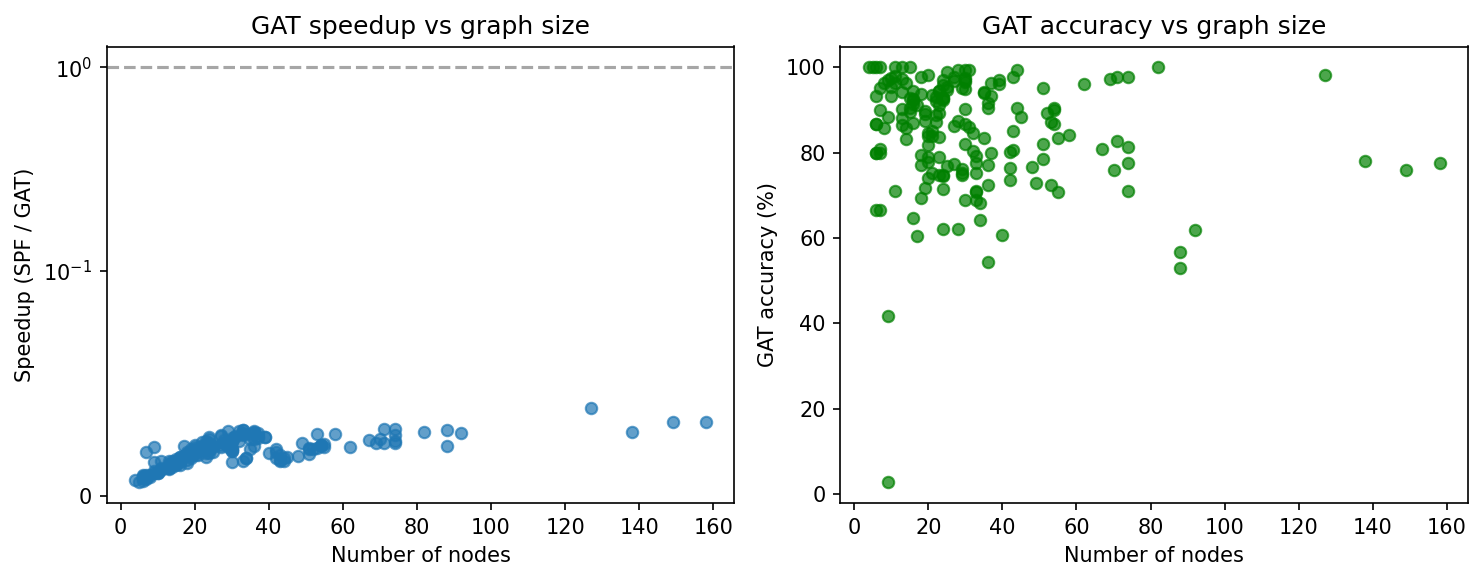}
    \caption{SPF vs.\ GAT inference speed and next-hop accuracy on 180 Internet Topology Zoo graphs.}
    \label{fig:spf_vs_gat}
\end{figure}

\section{Conclusion and Future Work}
We investigated whether a GAT can learn routing heuristics from synthetic graphs and transfer structural knowledge to unseen real-world networks to predict the most possible next hop on the shortest path. Our proposed GATNextHop model achieved 84.2\% accuracy and 84.6\% accuracy when only using betweenness as a node feature.

An ablation study reveals that betweenness centrality is the single most effective node-level feature that helps shortest-path routing's next-hop prediction, which aligns with the definition of betweenness centrality as a measurement of how often a node lies on shortest paths.
Benchmark comparison of our model against shows that in single-source queries, Dijkstra is still dominantly faster (50 times) compared to inference, although the trend as $n$ grows shows potential benefits of GNN in large $n$ and the amortized speedup of multiple queries and dynamic topologies are not explored in this paper.

Overall, the paper demonstrates the transferability and generalization potential of GAT for learning structural information from synthetic graphs and predicting the shortest path in real-world ISP networks.

We acknowledge that in a static setting where the graph structure is known and loaded to memory, Dijkstra remains the faster approach for shortest-path routing and that the overhead of GAT is not justified. Moreover, an $84\%$ accuracy means roughly 1 in 6 queries select suboptimal routes.}

Future directions include (i) simulating topologies in a SDN environment and experiment with topology changes or failure/fallbacks, (ii) incorporate graph-size augmentation and continuous learning techniques to adapt to larger networks, (iii) evaluate amortized per-query latency when GNN embeddings are precomputed once per topology and compare speedup with SPF via embedding caching, (iv) benchmark GAT against other GNN paradigms like GCN, GraphSAGE, KAN, GIN, and MPNN to isolate the contribution of attention mechanism in routing strategies, (v) explore broader applications in traffic engineering, load balancing, cloud and distributed environments, (vi) include and experiment with other graph-level features (eg. radius, average longest path) and compare the results.




\bibliographystyle{IEEEtran}
\bibliography{references}

\end{document}